\documentclass{article}
\usepackage{spconf,amsmath,graphicx}
\usepackage{graphicx}
\usepackage{textcomp}
\usepackage{xcolor}
\usepackage{algorithm}
\usepackage{algpseudocode}
\usepackage{amsmath}
\usepackage{url}
\usepackage{multirow}
\usepackage{setspace}

\newcommand{\probP}{\text{I\kern-0.15em P}}

\title{A few-shot learning approach with domain adaptation for personalized real-life stress detection in close relationships}

\name{Kexin Feng${^{*}}$, Jacqueline B. Duong${^{\dagger}}$, Kayla E. Carta${^{\dagger}}$, Sierra Walters{${^{\dagger}}$}, \vspace{-15pt}
\thanks{This research is based on work supported by NSF BCS-1627272 (Margolin, PI), NIH NIMH R42MH123368 (Timmons, PI), and NSF IIS-2046118 (Chaspari, PI). A. C. Timmons owns intellectual property and stock in Colliga Apps Corp. and could benefit financially from commercialization of related research. The code of this work is available at: \protect\url{https://github.com/HUBBS-Lab-TAMU/couple-stress-detection}.}}
\address{
\textit{Gayla Margolin${^{\mathsection}}$, Adela C. Timmons${^{\dagger}}$, Theodora Chaspari${^{*}}$} \vspace{5pt} \\ 
${^{*}}$Texas A\&M University, College Station, TX, USA \\
${^{\mathsection}}$University of Southern California, Los Angeles, California, USA \\
${^{\dagger}}$University of Texas, Austin, TX, USA \\
${^{*}}$\{kexin, chaspari\}@tamu.edu, ${^{\mathsection}}$margolin@usc.edu\\
${^{\dagger}}$\{jduong, kcarta, snwalters\}@utexas.edu, ${^{\dagger}}$adela.timmons@austin.utexas.edu}

\begin{document}

\maketitle

\begin{abstract}\vspace{-5pt}
We design a metric learning approach that aims to address computational challenges that yield from modeling human outcomes from ambulatory real-life data. The proposed metric learning is based on a Siamese neural network (SNN) that learns the relative difference between pairs of samples from a target user and non-target users, thus being able to address the scarcity of labelled data from the target. The SNN further minimizes the Wasserstein distance of the learned embeddings between target and non-target users, thus mitigating the distribution mismatch between the two. Finally, given the fact that the base rate of focal behaviors is different per user, the proposed method approximates the focal base rate based on labelled samples that lay closest to the target, based on which further minimizes the Wasserstein distance. Our method is exemplified for the purpose of hourly stress classification using real-life multimodal data from 72 dating couples. Results in few-shot and one-shot learning experiments indicate that proposed formulation benefits stress classification and can help mitigate the aforementioned challenges.
\end{abstract}
\vspace{-5pt}
\begin{keywords}
Few-shot learning, Siamese neural network, Wasserstein distance, stress detection
\end{keywords}
\vspace{-5pt}
\section{Introduction}
\vspace{-8pt}
\label{sec: intro}

The increasing availability of Internet of Things (IoT) technologies and wearable devices has enabled the monitoring of human states outside the lab resulting in the acquisition of real-life, multimodal, temporal data, that can serve as the foundation for automated algorithms for tracking an individual's internal and contextual states. Detecting (or predicting) one's changing state can contribute to behavioral support delivery by providing the right type and amount of feedback at the right time~\cite{nahum2018just}. Particularly, the prevalence of prolonged psychological stress globally~\cite{world2015world} renders such applications more relevant than ever. Beyond the individual level, ambulatory monitoring can be considered at an interpersonal context for detecting stress between interacting partners in close relationships (i.e., romantic couples, families)~\cite{timmons2017new}. In this context, ambulatory data can be used to monitor interconnected stress patterns between the partners, thus contributing to behavioral interventions that can demonstrate pathways for handling stress at the interpersonal level~\cite{ditzen2008positive,kuhn2017zooming}.


Designing machine learning (ML) models using real-world human-centered data presents unique computational challenges. First, labels obtained from the ambulatory data may be costly and tend to depict low temporal resolution~\cite{narayanan2013behavioral}. In particular, labels in ambulatory monitoring are commonly acquired via ecological momentary assessment (EMA), which is usually not administered exactly at the time of the stressor stimuli. While third-party annotation can potentially fix this issue, such type of annotation is difficult (or non-feasible) to obtain, since it requires human experts reviewing large portions of longitudinal data. Second, due to the inherent inter-individual variability, the distribution of features that result from ambulatory data may be different among individuals~\cite{gupta2020sub}. Due to this domain mismatch, the performance of the models may vary across participants. Third, the base rate of occurrence of the focal behaviors is usually unknown and may vary among users~\cite{spruijt2014dynamic}.

To address the above challenges, we propose a distance learning method implemented with a Siamese neural network (SNN) that learns stress embeddings between labelled samples from non-target participants and a few labelled samples from a target participant. We also integrate into the proposed approach a distance-based criterion based on the Wasserstein distance that aims to minimize the mismatch between the unlabelled samples of the target participant and the rest of the participants. Finally, to address the challenge of unknown stress base rate, we compute the Wasserstein distance across various hypothesized stress base rates for the target participant and minimize the loss function that yields the lowest value. We evaluate the proposed method on a multimodal dataset that includes ambulatory data from 72 dating couples performing daily activities in a day. Results demonstrate that minimizing the mismatch between labelled and unlabelled samples via the Wasserstein distance and leveraging base rate information can improve stress classification.



\vspace{-5pt}
\section{Previous work} \label{sec: previous}
\vspace{-8pt}
One of the first studies on real-world stress detection by Healey \textit{et al.} analyzed stress based on drivers' physiological signals\cite{healey2005detecting}. Using this data, Keshan \textit{et al.} applied ML to achieve the personalized detection of driver stress \cite{keshan2015machine}. Later on, Koldijk \textit{et al.} examined stress during cognitive performance tasks using multimodal data \cite{koldijk2014swell}. In terms of detecting stress via unobtrusive ambulatory sensors, Sagbas \textit{et al.} \cite{saugbacs2020stress} and Garg \textit{et al.} \cite{garg2021stress} leveraged smartphone and wearable devices to detect stress under predefined tasks and daily routine activities. Recent work has also investigated ML algorithms to detect stress based on unscripted real-life activities \cite{hovsepian2015cstress, sysoev2015noninvasive, gjoreski2016continuous}. The majority of this work is limited to a relatively low number of participants and usually relies on conventional ML algorithms. Limited prior work has examined deep learning methods for stress detection \cite{panicker2019survey,can2019stress}.

Metric learning is uniquely positioned to learn patterns from scarcely labelled data and effectively model the data distribution mismatch between different individuals \cite{yang2006distance,kulis2013metric}. It can effectively model the relative distance between samples from various classes, rather than the absolute distribution of each class, which typically requires a large number of labels. Thus, metric learning provides an elegant solution for modeling ``bouts" from the neutral state within an individual, thus potentially helping address the inherent inter-individual variability. It can be implemented via a SNN architecture that takes a pair of samples as input and identifies whether these two samples are from the same class \cite{chicco2021siamese}. Prior work in emotion recognition has demonstrated the effectiveness of SNN for transferring knowledge between domains \cite{zhang2022few,feng2021few}, rather than modeling inter-individual differences.



The contributions of this study are as follows: (1) In contrast to the majority of personalized stress detection algorithms that estimate absolute distributions of the neutral and stress classes, the proposed approach models the relative difference between the two, thus, potentially being able to effectively recognize ``bouts" from the neutral state with a small amount of labelled data; (2) Metric learning has been typically used to address mismatch between different recording conditions or behavior elicitation methods. Here, we propose an alternative formulation that models distribution mismatch between individuals; and (3) Our formulation estimates and subsequently integrates information regarding the base rate of the focal stress behaviors, which significantly varies between participants, for improved stress classification.

\vspace{-5pt}
\section{Proposed Methodology} \label{sec: method}
\vspace{-8pt}

\begin{figure}[!tb]
  \centering
  \includegraphics[width=0.9\linewidth, trim = 0.2cm 0.1cm 0.2cm 0.2cm, clip=true]{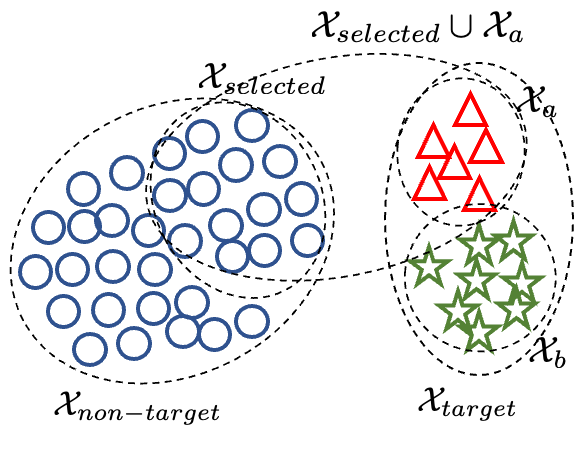}\vspace{-10pt}
 \caption{Schematic representation of target and non-target samples in the proposed few-shot learning formulation.}
\label{fig:wasserstein}
\vspace{-10pt}
\end{figure}

We formulate a metric learning algorithm that classifies a target participant's stress $y\in{0,1}$ based on a multimodal input $\mathbf{x}$. We leverage all labelled samples $\mathcal{X}_{non-target},\mathcal{Y}_{non-target}$ from the non-target participants and few labelled samples $\mathcal{X}_a,\mathcal{Y}_a$ from the target participant, where $\mathcal{X}_{non-target}$ and $\mathcal{X}_a$ are the multimodal feature matrices and $\mathcal{Y}_{non-target}$ and $\mathcal{Y}_a$ are the binary stress labels, and aim to estimate stress for the unlabelled samples $\mathcal{X}_b$ of the target. Grounded in prior work that suggests it is beneficial to leverage the knowledge from a subset of the participants closest to the target one \cite{zhao2022personalized}, we select the $n$ samples from the target participant $\{\mathbf{x_i},y_i\}\in \{\mathcal{X}_{non-target},\mathcal{Y}_{non-target}\}$ that depict the smallest $l2$-norm distance to the samples from the non-target participant $\{\mathbf{x_j},y_j\} \in \{\mathcal{X}_{a},\mathcal{Y}_a\}$, where $y_i=y_j$. These are denoted as $\mathcal{X}_{selected}\subset\mathcal{X}_{non-target}$ (Fig.~\ref{fig:wasserstein}).

To address the first challenge of limited labelled data from the target user, we adopt a SNN that learns the similarity between pairs of samples $(\mathbf{x_i},\mathbf{x_j})$ from the target user, $\mathbf{x_j}\in\mathcal{X}_a$ and $y_j\in\mathcal{Y}_a$, and non-target users, $\mathbf{x_i}\in\mathcal{X}_{selected}$ and $y_i\in\mathcal{Y}_{selected}$. The SNN contains two identical sub-networks represented by fully connected layers that conduct the same non-linear transformation $f_\mathbf{V}$ with weight parameters $\mathbf{V}$. We assign to each pair a binary similarity label, i.e., 0, if $y_i=y_j$; or 1, if $y_i\neq y_j$, which serves as the output of the SNN. During training, we minimize the cross entropy loss that represents whether the input samples belong to the same class:
\begin{equation}
\resizebox{0.5\textwidth}{!}{$
\begin{aligned}
C(\mathbf{V}) = -\frac{1}{N_s}\sum_{\substack{\mathbf{x_i}\in\mathcal{X}_{selected}\\y_i\in\mathcal{Y}_{selected}\\ \mathbf{x_j}\in\mathcal{X}_{a}, \; y_j\in\mathcal{Y}_{a},\; y_i=y_j}}\log\left(\|f_\mathbf{V}\left(\mathbf{x_i}\right)-f_\mathbf{V}\left(\mathbf{x_j}\right)\|\right)\\
-\frac{1}{N_s}\sum_{\substack{\mathbf{x_i}\in\mathcal{X}_{selected}\\y_i\in\mathcal{Y}_{selected}\\ \mathbf{x_j}\in\mathcal{X}_{a}, \; y_j\in\mathcal{Y}_{a},\; y_i=y_j}}\log\left(1-\|f_\mathbf{V}\left(\mathbf{x_i}\right)-f_\mathbf{V}\left(\mathbf{x_j}\right)\|\right)
\end{aligned}$
}
\label{eq:1}
\end{equation}
where $N_s=2|\mathcal{X}_{selected}|$.
We address the second challenge of domain mismatch between the target and non-target participants by minimizing the relative difference between the embedding of unlabelled samples of the target user and embedding of the labelled samples of the non-target users. To achieve this, we minimize the Wasserstein distance $W(\cdot)$ between the probabilistic distribution $\probP_{f_\mathbf{W}}^{\mathcal{X}_a\cup\mathcal{X}_{selected}}$ of selected samples from the non-target participants, $\mathcal{X}_{selected}$, and labelled samples from the target participant, $\mathcal{X}_a$, and the probabilistic distribution $\probP_{f_\mathbf{W}}^{X_b}$ of the unlabelled samples from the target participant $\mathcal{X}_b$:
\begin{equation}
\resizebox{0.49\textwidth}{!}{$
\begin{aligned}
    D(\mathbf{V})=&W(\probP_{f_\mathbf{W}}^{\mathcal{X}_a\cup\mathcal{X}_{selected}},\probP_{f_\mathbf{W}}^{X_b})\\
    &=\frac{1}{N_b}\inf_{\gamma\in\mathcal{J}(\probP_{f_\mathbf{W}}^{\mathcal{X}_a\cup\mathcal{X}_{selected}},\probP_{f_\mathbf{W}}^{X_b})}\int{\|X_1-X_2\|d\gamma(X_1,X_2)}
\end{aligned}$
}
\label{eq:2}
\end{equation}
where $\mathcal{J}(\probP_{f_\mathbf{W}}^{\mathcal{X}_a\cup\mathcal{X}_{selected}},\probP_{f_\mathbf{W}}^{X_b})$ and $N_b=|\mathcal{X}_b|$ denotes all joint distributions. In practice, the Wasserstein distance is approximated using Sinkhorn iterations between $\mathcal{X}_{b}$ and $N_b$ samples randomly drawn from $(\mathcal{X}_{a} \cup \mathcal{X}_{selected})$.

To tackle the third challenge regarding the unknown base rate of stress instances, we randomly select samples drawn from $(\mathcal{X}_{a} \cup \mathcal{X}_{selected})$ with different stress base rates, and calculate the Wasserstein distance $D_p(\mathbf{V})$ between $(\mathcal{X}_{a} \cup \mathcal{X}_{selected})$ and $\mathcal{X}_b$ for each base rate $p$. The samples selected based on the minimum Wasserstein distance $D^*_p(\mathbf{V})=\min_p{D_p}(\mathbf{V})$ will be the ones that will be used when calculating (\ref{eq:2}). We consider three different base rates: (1) low stress ($p=0.05$); (2) balanced ($p=0.5$); and (3) high stress ($p=0.95$). 

Combining the above, the weights $\mathbf{C}$ of the SNN are learned so that they minimize the following loss function:
\begin{equation}
    \mathbf{V}^*=\min_{\mathbf{V}}\left(C(\mathbf{V})+\lambda D^*_p(\mathbf{V})\right)
\label{eq:3}
\end{equation}

\noindent After we learn the feature embeddings $f_\mathbf{V}$ based on (\ref{eq:3}), we train a Support Vector Machine (SVM) on stress classification based on these embeddings. The training and test data include samples from $\mathcal{X}_{selected}$ and $\mathcal{X}_b$, respectively.


\begin{figure}[!tb]
  \centering
  \includegraphics[width=1\linewidth, trim = 0.2cm 0.1cm 0.25cm 0.2cm, clip=true, scale=1]{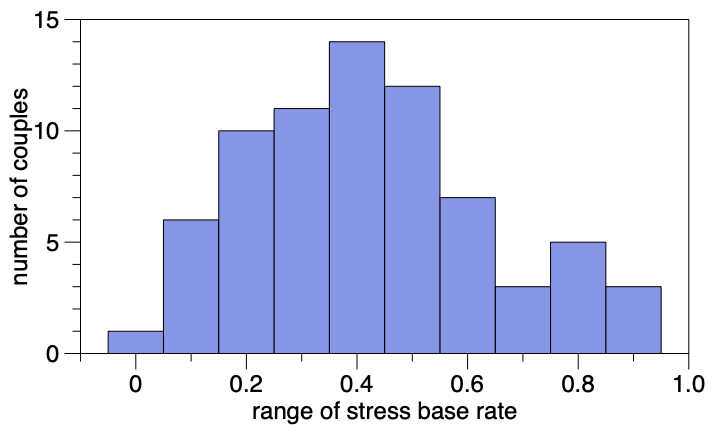}\vspace{-10pt}
 \caption{The distributions of the stress base rate among couples.}
\label{fig: stress base rate}
\vspace{-5pt}
\end{figure}

\begin{figure*}[!tb]
  \begin{minipage}{0.49\linewidth}
  \centering\includegraphics[trim = 0.5cm 0.3cm 1.6cm 1.4cm, clip=true, scale=0.59]{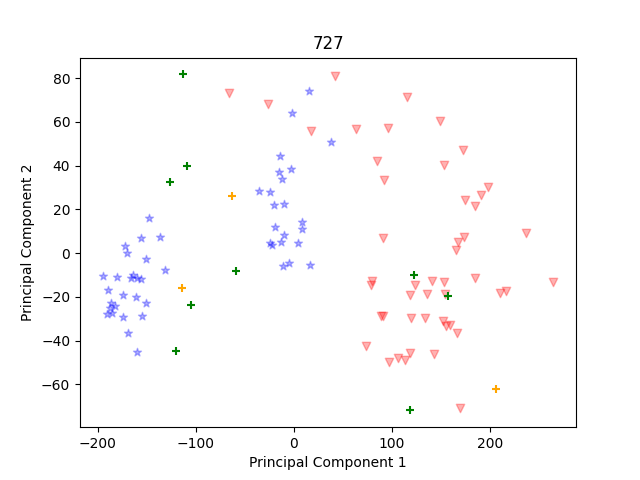}
  \centerline{(a) Baseline 1}\medskip
  \end{minipage}
  \begin{minipage}{0.49\linewidth}
 \centering\includegraphics[trim = 0.5cm 0.3cm 1.6cm 1.4cm, clip=true, scale=0.59]{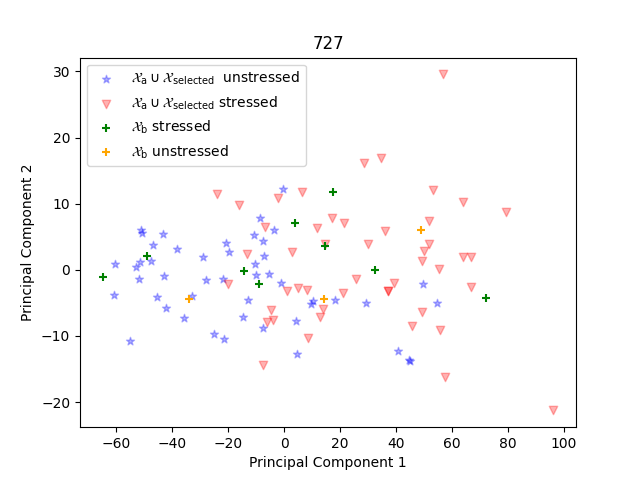}
  \centerline{(b) Proposed}\medskip
  \end{minipage}
  \hfill
  \vspace{-10pt}
 \caption{An example of the first two Principal Component Analysis (PCA) dimensions resulting from the learned embeddings for the baseline and proposed methods. While the models are trained using all samples from the set $\mathcal{X}_{a} \cup \mathcal{X}_{selected}$, only a random set of 96 samples is plotted for visualization purposes. The legend in (b) applies to both plots.}
\label{fig: pca}
\vspace{-10pt}
\end{figure*}

\vspace{-5pt}
\section{Experiments} 
\label{sec: experiment}
\vspace{-8pt}

\subsection{Data description} \label{ssec: data description}
\vspace{-8pt}
Our dataset is collected as part of the University of Southern California (USC) Couple Mobile Sensing Project\footnote{\url{ https://colliga.io/data-repository/}} introduced by Timmons {\it et al.} \cite{timmons2017using}. It includes 109 couples fluent in English who have been dating for at least 2 months (average: 32.2 months). The range of ages is between 18 and 25, with an average of 23.1 and a standard deviation of 3.0. This dataset also has a great variety of races (23.9\% Hispanic, 16.1\% African American, 12.8\% Asian, 15.6\% Multi-racial, 3.7\% self-identified as other, and 0.5\% Native Hawaiian). The couples are equipped with a smartphone and two wearable devices, and were asked to spend their day as they would normally do. They completed an ecologocal momentary assessment (EMA) inquiring about their level of stress reported at a 0-100 scale every hour. The smartphone records a 3-minute audio per 12 minutes. Multimodal measures are extracted within each hour similar to \cite{timmons2017using}. The 10 acoustic measures include the mean, median, standard deviation, and max values of pitch and loudness, and the min and range of pitch. Our linguistic feature is 73 dimensions extracted using the Linguistic Inquiry and Word Count (LIWC) lexicon, including 28 linguistic factors (e.g., pronouns), 33 psychological constructs (e.g., emotions), 6 personal concern categories (e.g., money), and 6 paralinguistic variables (e.g., fillers). \cite{pennebaker2015development}. The 8 physiological measures include the mean skin conductance level (SCL), body temperature, interbeat interval (IBI), heart rate variability (HRV), beats per minute (BPM), and the frequency of skin conductance responses (SCRs). This resulted in a 91-dimensional feature vector.
After discarding the samples/hours with missing features or labels, we select the couples that reported at least once stressed and unstressed in the first 40\% of their data. A total of 72 couples satisfy this criterion and are used in this study. This includes a total of 763 unstressed samples (hours) and 557 stressed samples (hours).



\vspace{-5pt}
\subsection{Experimental setting and baseline method} \label{ssec: baseline}
\vspace{-8pt}
Each SNN sub-network includes an input layer with 91 units, two hidden layers with 32 and 16 units, respectively, followed by LeakyReLU activation, and one output layer with 1 unit and Sigmoid activation. The SNN is trained using the Adam optimizer with a learning rate of 0.001. During the pairing process, for every sample $\mathbf{x_i}\in\mathcal{X}_{selected}$, we randomly select two samples from $\mathbf{x_j}\in\mathcal{X}_a$, and thus, form two pairs that either belong to the same or different class. The $\lambda$ in Equation \ref{eq:3} is set to 0.15. 
Baseline 1 is a vanilla SNN that solely compares the relative distance between pairs of samples from the target and non-target speakers, minimizing the loss $C(\mathbf{V})$ of (\ref{eq:1}). Baseline 2 aims to minimize the $C(\mathbf{V})+\lambda D(\mathbf{V})$ of (\ref{eq:2}), in which the Wasserstein distance is incorporated but without addressing the unknown base rate challenge.

Experiments are conducted in a leave-one-couple-out cross-validation, based on which data from a couple are included in the test, while data from the remaining couples are in the train. This procedure is repeated as many times as the total number of couples. For the few-shot learning evaluation, we set $\mathcal{X}_{a}$ to be the first 40\% of the $\mathcal{X}_{target}$, since in clinical settings labels from the first stages of the study will become available first. For the one-shot learning, the $\mathcal{X}_{a}$ comprises of the first labelled sample from the target couple. Due to the inherent difference between the two settings, $|\mathcal{X}_{selected}|=256$ and $|\mathcal{X}_{selected}|=128$, while the number of epochs is 10 and 3 for the few-shot learning and one-shot learning, respectively. The F1-score computed for stress and no-stress samples is reported as an evaluation metric, as well as the macro-F1-score for both classes.

\begin{table}[!tb]
\caption{F1-scores (\%) of baselines and proposed methods using one-shot and few-shot learning.}
\scriptsize
\begin{tabular*}{0.48\textwidth}{@{\extracolsep{\fill}}lllll}
\hline
 & Method \& Loss Function & Stress & No-stress & Macro \\ \hline
\multirow{3}{*}{One-shot} & Baseline 1: $C(\mathbf{V})$ & 48.0 & 52.2 & 50.1 \\
 & Baseline 2: $C(\mathbf{V})+\lambda D(\mathbf{V})$  & 47.2 & 55.3 & 51.3 \\
 & Proposed: $C(\mathbf{V})+\lambda D^*_p(\mathbf{V})$ & 50.3 & 56.4 & 53.3 \\ \hline
\multirow{3}{*}{Few-shot} & Baseline 1: $C(\mathbf{V})$ & 48.8 & 61.8 & 55.3 \\
 & Baseline 2: $C(\mathbf{V})+\lambda D(\mathbf{V})$ & 49.4 & 62.0 & 55.7 \\
 & Proposed: $C(\mathbf{V})+\lambda D^*_p(\mathbf{V})$ & 55.6 & 65.7 & 60.7 \\ \hline
\end{tabular*} \label{tab: final result}
\vspace{-10pt}
\end{table}

\vspace{-5pt}
\section{Results } 
\vspace{-8pt}
To empirically illustrate the inherent domain mismatch across participants in the data, we train a vanilla feedforward NN using a leave-one-couple-out cross-validation. This model has a similar structure to the SNN. The macro-F1 score of the cross-validation is 0.473. This score is slightly worse than random guessing, suggesting a large mismatch between target and non-target participants. We further plot the distribution of stress base rate for each participant in Fig. \ref{fig: stress base rate}. Even though the median stress base rate is around 0.3, there are many couples who depict very low or very high stress base rate.

The proposed approach for mitigating domain mismatch between the target and non-target participants depicts improved results compared to both baseline methods (Table \ref{tab: final result}). Solely modeling the relative distance between target and non-target (Baseline 1) yields the lowest results. The unsupervised domain adaptation without considering the stress base rate (Baseline 2) provides marginal benefits compared to Baseline 1. The one-shot learning yields overall a low performance indicating the difficulty of the task. To further better understand the impact of domain adaptation, we plot the first two Principal Components from the learned embeddings and show an example in Fig. \ref{fig: pca}. Baseline 1 learns a feature embedding that projects all test samples $\mathcal{X}_{b}$ from the target couple mostly close to the no-stress samples from the non-target participants. In contrast, the proposed method achieves better differentiation between stress and no-stress samples from the target participant, which coincides with the stress and no-stress subspaces of the non-target participants.


\vspace{-5pt}
\section{Conclusions}
\vspace{-8pt}
We propose a metric learning method implemented with a SNN that aims to address the small number of labelled samples from a target user, the inherent domain mismatch between users, and unknown base rates of focal behaviors. Our method provides a 5.3\% absolute performance increase compared to the vanilla few-shot learning method. As part of our future work, we plan to leverage additional participant-dependent information, such as the psychological trait characteristics of each participant, to achieve further personalization. We further plan to explore additional base rate estimation methods based on the labelled target data. Finally, we would like to explore additional types of distances to more effectively model the domain mismatch between participants.


\renewcommand{\normalsize}{\fontsize{9.5}{11.00}\selectfont}
\normalsize
\vfill\pagebreak
\let\oldbibliography\thebibliography
\renewcommand{\thebibliography}[1]{%
  \oldbibliography{#1}%
  \setlength{\itemsep}{-2pt}%
}

\bibliographystyle{IEEEbib}
\bibliography{refs}

\end{document}